\documentclass[twoside]{article}
\usepackage[dvipsnames]{xcolor}
\usepackage{graphicx}
\usepackage{amsthm}
\usepackage{amsmath}
\usepackage{mathtools}
\usepackage{mathrsfs}
\usepackage{amssymb}
\usepackage{lipsum}
\usepackage{etoolbox}
\usepackage[linesnumbered,lined,boxed,ruled]{algorithm2e}
\usepackage[section]{placeins}
\usepackage[breaklinks,colorlinks,citecolor=blue,linkcolor=blue,urlcolor=blue]{hyperref}
\usepackage[all]{hypcap}
\usepackage{cancel}
\usepackage{csquotes}
\usepackage[toc,page]{appendix}
\usepackage{multirow}
\usepackage{datetime}
\usepackage{times}
\usepackage[format=plain, labelfont={bf,it}, textfont=it]{caption}
\usepackage{tcolorbox}
\usepackage{natbib, bibentry}
\usepackage{footnote}
\usepackage{orcidlink}
\usepackage{fancyhdr}

\DeclarePairedDelimiter{\norm}{\lVert}{\rVert} 

\usepackage{scalerel}
\usepackage{stackengine,wasysym}

\newdateformat{mdydate}{\monthname[\THEMONTH] \THEDAY, \THEYEAR}
\newdateformat{monthyeardate}{\monthname[\THEMONTH] \THEYEAR}
\newdateformat{vmonthyeardate}{\THEYEAR.\THEMONTH.\THEDAY}

\newcommand{\be}{\begin{equation}}
\newcommand{\ee}{\end{equation}}
\newcommand{\bea}{\begin{eqnarray}}
\newcommand{\eea}{\end{eqnarray}}

\usepackage{geometry}
\graphicspath{{figures}}

\newcommand{\volume}{2}
\newcommand{\firstpage}{1}
\newcommand{\lastpage}{14}
\newcommand{\yyyy}{2025}
\newcommand{\mm}{January}
\newcommand{\dd}{27}
\newcommand{\authors}{Kruijssen, Valieva \& Longmore}
\newcommand{\fulltitle}{Optimizing Decentralized Online Learning for\\[0.5ex] Supervised Regression and Classification Problems}
\newcommand{\shorttitle}{Optimizing Decentralized Online Learning for Regression and Classification}

\newcommand{\doi}{10.70235/allora.0x\volume\ifnum\numexpr\firstpage<10 000\else\ifnum\numexpr\firstpage<100 00\else\ifnum\numexpr\firstpage<1000 0\fi\fi\fi\firstpage}
\fancypagestyle{firstpage}{%
    \fancyhead[L]{\footnotesize Allora Decentralized Intelligence \textbf{\volume}, \firstpage--\lastpage; \yyyy\ \mm\ \dd}
    \fancyhead[R]{\footnotesize doi:\href{https://doi.org/\doi}{\textcolor{blue}{\doi}}}
    \fancyfoot{}
    
}

\AtBeginDocument{\thispagestyle{firstpage}}

\begin{document}

\title{\fulltitle}
\author{\authors}
\date{\monthyeardate{\today}}

\vskip30mm
\begin{center}
\begin{minipage}{170mm}
\begin{center}
\vskip5mm
{\fontsize{15pt}{15pt}\textbf{\fulltitle}}
\vskip5mm
J.~M.~Diederik Kruijssen\,${\orcidlink{0000-0002-8804-0212}\,^1}$,
Renata Valieva\,${\orcidlink{0000-0002-7256-5321}\,^1}$ \&
Steven~N.~Longmore\,${\orcidlink{0000-0001-6353-0170}\,^{1,2}}$
\vskip1mm
$^1$\textit{Allora Foundation},
$^2$\textit{Liverpool John Moores University}
\end{center}
\end{minipage}
\end{center}
\vspace{3mm}

\begin{abstract}
\noindent Decentralized learning networks aim to synthesize a single network inference from a set of raw inferences provided by multiple participants. To determine the combined inference, these networks must adopt a mapping from historical participant performance to weights, and to appropriately incentivize contributions they must adopt a mapping from performance to fair rewards. Despite the increased prevalence of decentralized learning networks, there exists no systematic study that performs a calibration of the associated free parameters. Here we present an optimization framework for key parameters governing decentralized online learning in supervised regression and classification problems. These parameters include the slope of the mapping between historical performance and participant weight, the timeframe for performance evaluation, and the slope of the mapping between performance and rewards. These parameters are optimized using a suite of numerical experiments that mimic the design of the Allora Network, but have been extended to handle classification tasks in addition to regression tasks. This setup enables a comparative analysis of parameter tuning and network performance optimization (loss minimization) across both problem types. We demonstrate how the optimal performance-weight mapping, performance timeframe, and performance-reward mapping vary with network composition and problem type. Our findings provide valuable insights for the optimization of decentralized learning protocols, and we discuss how these results can be generalized to optimize any inference synthesis-based, decentralized AI network.
\end{abstract}
\vspace{3mm}

\section{Introduction} \label{sec:intro}
Decentralized learning is a key ingredient of blockchain-based artificial intelligence (AI) networks, which enable the secure coordination of data owners, data processors, AI models, and the end users utilizing these insights \citep[e.g.][]{dinh18,wang21,qammar23,bhumichai24}. The trustless and permissionless nature of blockchains is ideally suited for coordinating independent actors in a collaboration game \citep[e.g.][]{nakamoto08,buterin14}. Over the past years, several examples of on-chain decentralized AI have been introduced, either in the form of on-chain models \citep[e.g.][]{harris19,zheng24} or in the form of model coordination networks \citep[e.g.][]{craib17,rao21,kruijssen24}. Any such network requires a mechanism to combine the raw inferences of individual participants into a single network inference (referred to as Inference Synthesis), and to correctly incentivize these participants by rewarding them for their contributions to this network inference.

The Inference Synthesis mechanisms and incentive structures that are used by these networks have a number of free parameters that can be tuned to optimize the performance of the network. Despite the general applicability of such parameter calibrations and the numerous decentralized AI networks that have recently appeared, there exists no systematic study that performs such a calibration. This work aims to fill this gap by systematically optimizing a set of generalizable parameters governing Inference Synthesis and reward distribution in decentralized AI networks. The heterogeneity of the problem types and solution architectures is considerable, and therefore we reduce the problem to a small set of essential elements.

Any decentralized learning protocol that combines multiple model outputs into a single, joint inference must decide on a mapping between the historical performance of a participant and their weight in determining the network inference. Some protocols perform this calculation on-chain \citep[e.g.][]{kruijssen24}, whereas others outsource this in the form of a voting contest among network participants \citep[e.g.][]{craib17,rao21}. Regardless of how the performance-weight mapping is consolidated, the simplest way in which such a mapping can be characterized is by its slope, i.e.\ how quickly the weight of a participant increases or decreases as their performance changes. The second free parameter is to quantify the word `historical' in `historical performance', i.e.\ over what timeframe the performance is considered and how strongly the most recent performance is weighted relative to the historical performance. Moving averages are a key tool to quantify this property, and the particular form of the moving average (e.g.\ simple, exponential, or something else) is a free parameter that can be tuned to optimize the performance and adaptability of the network. Finally, decentralized learning protocols must decide how to distribute rewards among the participating agents. This is often done based on the participant's performance (e.g.\ score) or their commitment to the network (e.g.\ stake). Similarly to the mapping between performance and weight, the simplest form of characterizing this mapping between performance and rewards is by calibrating the slope of the mapping, i.e.\ how quickly the reward of a participant increases or decreases as their performance changes.

These forms of parameter calibration are a requirement for any decentralized learning protocol that synthesizes a network inference from a large pool of raw inferences, independently of the specific decentralized learning protocol used. In this paper, we perform parameter optimization using an experimental setup that mimics the design of the Allora Network \citep{kruijssen24}, due to its transparent design and tractable parameter functionality. Specifically, we focus on parameters that describe the aforementioned mappings between (historical) performance and weight, and performance and rewards, as a function of the network composition (i.e.\ the number of participants) and the problem type. While Allora's design was initially tailored to regression tasks, we extend this design to additionally cover classification tasks. This enables the comparison of parameter optimization for both problem types.

The structure of this paper is as follows. In \S\ref{sec:method}, we describe the decentralized learning framework of the Allora Network and introduce the key concepts that are relevant to the present work, summarizing the regression setup from \citet{kruijssen24} in \S\ref{sec:method_regr}, presenting the extension of the Allora Network to handle classification tasks in \S\ref{sec:method_class}, and summarizing the incentive structure in \S\ref{sec:incentive}. In \S\ref{sec:ics}, we describe the synthetic data generation process that is used in our numerical experiments for both regression and classification tasks. In \S\ref{sec:opt}, we describe the parameter optimization process and present the results, showing how the optimal performance-weight mapping and performance-reward mapping depend on the network composition and problem type. Finally, we summarize and discuss the results in \S\ref{sec:disc}.

\section{Decentralized Learning as Facilitated by the Allora Network} \label{sec:method}

\subsection{Supervised Regression} \label{sec:method_regr}
We carry out the experiments described in this work using a simulator that describes the decentralized learning infrastructure of the Allora Network \citep{kruijssen24}. Allora is a state-of-the-art protocol that uses decentralized AI and ML to build, extract, and deploy AI predictions or inferences among its participants. It offers a formalized way to obtain the output of ML models in blockchain networks of virtual machines (VMs) and to reward the operators of AI nodes who create these inferences. In this way, Allora bridges the information gap between data owners, data processors, AI models, and the end users or consumers who have the means to execute on these insights. We refer to \citet[\S3.1]{kruijssen24} for details on the design of the network Inference Synthesis mechanism and its incentive structure. Key elements relevant to the present work are summarized below.

Allora is composed of `topics', which are sub-networks within which network participants collaborate to generate inferences and earn rewards. Each topic contains a short rule set that governs the interaction between the topic participants, including the target variable and the loss function that needs to be optimized by the topic network (where lower losses indicate better performance). Within the scope of a topic, network participants collaborate to generate a network inference for the target variable defined within the topic.

To provide a network inference for a given target variable, Allora synthesizes the combines from a set of $N_{\rm w}$ workers, which carry out two tasks. First, a worker can use its own ML model and private data set to generate a raw inference for the target variable of the topic, using its own data set $D_{ij}$ and model $M_{ij}$:
\be
\label{eq:inf_inference}
I_{ij} = M_{ij} (D_{ij}) .
\ee
Secondly, a worker can use another ML model and an augmented private data set to generate forecasted losses for both its own raw inference (if any) and the raw inferences of other workers. Formally, we express this such that a worker $k\in \{1,\ldots,N_{\rm f}\}$ produces an inference for the logarithm of the forecasted loss $L_{ijk}$ of the inference $I_{ij}$ produced by worker $j$, using its own augmented data set $D_{ijk}$ and model $M_{ijk}$:
\be
\label{eq:fore_inference}
\log{L_{ijk}} = M_{ijk}(D_{ijk}) ,
\ee
The network then uses each set of forecasted losses provided by an individual worker to define weights for the raw inferences and generate a corresponding forecast-implied inference. This form of feedback on the expected performance of workers under the current conditions is what creates Allora's unique context awareness, and ensures that Allora's network inference represents the optimal combination of raw and forecast-implied inferences. A second class of network participants, called reputers, then compares the performance of all inferences to the ground truth whenever it becomes available to provide the actual losses. These losses are used by the network to calculate a regret, which in turn is mapped through a potential function to define the weight corresponding to each raw or forecast-implied inference.

\subsection{Supervised Classification} \label{sec:method_class}
The network design described by \citet{kruijssen24} requires modification to be applicable to classification problems, where the network needs to infer the label corresponding to a data point. From a design perspective, the required changes are comparatively minor, and they are described here. In extending Allora to handle classification problems, we focus on the general form of multi-class classification, of which we consider binary classification to be a subset. We make no further assumptions with regards to the label set, i.e.\ it may be unbounded in principle, and the design allows the network to handle time-variable label sets. These decisions maximize the general applicability of our solution.

When dealing with a classification problem, we generalize the infrastructure described in \citet[\S3.1]{kruijssen24} to handle a vector of probabilities describing the likelihood of each possible label occurring. The ground truth used by reputers is then the single label that occurs in reality, and the adopted loss function must accommodate the hybrid nature of the inferred probability vector and the categorical ground truth, while also reducing the vector of individual loss contributions to a single loss per inference. Examples of loss functions that fulfil these conditions either natively or after modification include the categorical cross-entropy loss, the multi-class hinge loss \citep[e.g.][]{dogan16}, focal loss \citep{lin17}, \citet{kullback51} divergence, as well as simple modifications of the mean squared or absolute errors.

The raw inferences generated by workers are modified slightly relative to \autoref{eq:inf_inference} by acknowledging the vector form of the inferred probabilities:
\be
\label{eq:inf_inference_class}
\vec{I}_{ij} = M_{ij} (D_{ij}) .
\ee
In order to maximize the network's applicability and flexibility, the label set need not be specified upon topic creation, but may be variable and is dynamically defined during each epoch. Each worker contributes an inference as in \autoref{eq:inf_inference_class}, where each inference is a set of labels and label probabilities. The workers follow a label syntax defined at the topic level, so that the network can reduce the labels to the unique set reported at least once by all workers. Any labels that are missing in a worker's inference but do appear in other workers' inferences are assigned a zero probability for that worker. This way, the probability vector is dynamically defined each epoch, and the Inference Synthesis mechanism can be used to solve problems with unbounded or evolving label sets (examples range from well-scoped problems like chess move prediction or medical diagnosis to broader problem sets like object detection, product classification, named entity recognition, security monitoring and threat detection, and many others). Additionally, the topic creator can set a flag that requires inference workers to report $\lambda(\vec{I}_{ij})$, where $\lambda$ is the inverse of the sigmoid function (i.e.\ the logit function). This is particularly useful for classification problems that require high-precision inference for rare events.

By contrast to \autoref{eq:inf_inference_class}, the definition of the forecasted loss in \autoref{eq:fore_inference} is unchanged, as the loss function must return a single scalar value. The forecast-implied inference follows in the same way as for regression:
\be
\label{eq:fore_implied_inference_class}
\vec{I}_{ik} = \frac{\sum_j w_{ijk}\vec{I}_{ij}}{\sum_j w_{ijk}} ,
\ee
where the weights are a function of the forecasted losses, $w_{ijk}(L_{ijk})$, as described in eqs.~4{-}8 of \citet{kruijssen24}.

The network inference is a weighted average of the raw and forecast-implied inferences. Historical regrets are mapped through a potential function to define the weights, which are updated each time a ground truth becomes available and the reputers evaluate the loss function. As such, the network inference for classification tasks combines the inferences $\vec{I}_{ij}$ and the forecast-implied inferences $\vec{I}_{ik}$ similarly to \autoref{eq:fore_implied_inference_class}. We first define a new variable $\vec{I}_{il}$ that appends both sets of inferences in a new array with $l\in \{1,\ldots,N_{\rm i}+N_{\rm f}\}$. The network inference is then defined as
\be
\label{eq:net_inference}
\vec{I}_{i} = \frac{\sum_l w_{il}\vec{I}_{il}}{\sum_l w_{il}} .
\ee
All other Inference Synthesis steps described in \citet[\S3.1]{kruijssen24} remain unchanged, except for the substitution of inferred probability vectors for scalar inferences.

Confidence intervals (CIs) are generated by calculating weighted percentiles of the raw and forecast-implied inferences as described in \citet[\S3.2]{kruijssen24}. While this is a natural approach for obtaining CIs on the continuous target variables of regression problems, it requires minor modification for classification problems. For the vector target variables of classification problems, we perform the CI calculation independently on the individual label probabilities, and disregard any covariance between the labels. The CIs are based on percentiles $P$, which are chosen to be $P \in \{2.28, 15.87, 84.13, 97.72\}\%$ (with corresponding quantiles $q \equiv P / 100$) to mimic the 1$\sigma$ and 2$\sigma$ limits of a Gaussian distribution, even if there is no guarantee that the distribution of worker inferences is Gaussian.

We use a weighted cumulative distribution function (CDF) to interpolate and obtain the confidence intervals. However, the application of a weighted average to calculate the network inference in \autoref{eq:net_inference} implies that the variance of the network inference is smaller than the variance across the sample of worker inferences. To account for this variance reduction, we adjust the worker inferences before constructing the CDF. Specifically, we scale the deviations of each worker inference from the weighted mean by a factor of $1/\sqrt{N}$. The adjusted worker inference $I_{ilc}'$ for label $c$ is computed as
\be
\label{eq:adj_inference} 
I_{ilc}' = I_{ic} + \frac{I_{ilc} - I_{ic}}{\sqrt{N}} , 
\ee
where $I_{ic}$ represents the individual elements of the network inference from \autoref{eq:net_inference}. This adjustment reflects the decrease in the standard error of the mean with increasing sample size, ensuring that the confidence intervals appropriately reflect the uncertainty in the network inference.

For each label $c$, the inferences ${I}_{ilc}'$ and their corresponding weights $w_{il}\rightarrow w_{ilc}$ are first sorted over dimension $l$ such that $I_{ilc}'\leq I_{i,l+1,c}'$. For each label $c$ and each sorted worker inference $l$, the cumulative sum of the weights is calculated as
\be
\label{eq:cumsum}
c_{ilc} = \sum_{l'=1}^{l} w_{il'c} .
\ee
This cumulative sum $c_{ilc}$ carries a label subscript $c$, because the sorting operation on $I_{ilc}'$ and $w_{il}\rightarrow w_{ilc}$ (where $w_{ilc}$ is the sorted version of $w_{il}$ for label $c$) is performed separately for each label. The cumulative weights are then normalized to create a weighted cumulative distribution function (CDF):
\be
\label{eq:wcdf}
C_{ilc} = \frac{c_{ilc}-0.5w_{ilc}}{\max_l c_{ilc}} ,
\ee
where $\max_l c_{ilc}$ is the total sum of the weights. The term $0.5w_{ilc}$ is subtracted to locate the quantile value at the middle of the weight associated with $I_{ilc}'$.  We obtain the inference value $I^q_{ic}$ corresponding to the quantile $q$ by linear interpolation of the sorted list of data points using the weighted CDF. We define $l = \ell$ as the largest index where $C_{ilc} \leq q$ and calculate $I^q_{ic}$ as
\be
\label{eq:ci_class}
I^q_{ic} = \sigma\left\{\lambda(I_{i\ell c}') + \frac{q-C_{i\ell c}}{C_{i,\ell+1,c}-C_{i\ell c}} \left[\lambda(I_{i,\ell+1,c}')-\lambda(I_{i\ell c}')\right]\right\} ,
\ee
where $\sigma$ is the sigmoid function and $\lambda$ is its inverse (i.e.\ the logit function). The interpolation is performed in logit space to accommodate the bounded nature of the probability vectors within the domain $[0,1]$.

\subsection{Main Free Parameters of the Inference Synthesis Process}
The differing characteristics of loss functions used to evaluate regression and classification problems imply that key parameters of the Inference Synthesis process should be optimized separately for both cases. Specifically, we consider the potential function that is used to map regrets to weights, and to map worker scores to rewards (see \S3.1 and \S4.1 of \citealt{kruijssen24}). The potential function is defined as:
\be
\label{eq:potential}
\phi_{p,c}(x) = \ln{\left[1+{\rm e}^{p(x-c)}\right]} ,
\ee
and its gradient as:
\be
\label{eq:potential_grad}
\phi_{p,c}'(x) = \frac{p}{{\rm e}^{-p(x-c)}+1} .
\ee
At any given epoch, regrets are mapped to weights by normalizing them by their standard deviation and passing them through the potential function gradient of \autoref{eq:potential_grad}. Analogously, the scores assigned to each individual worker are normalized by their standard deviation and passed through the potential function of \autoref{eq:potential} to obtain the reward fraction assigned to each worker. For the regression problems discussed in \citet{kruijssen24}, we define a fiducial value of $p=3$ at a fixed $c=0.75$. In this paper, we optimize the value of $p$ independently for the regret-weight mapping and the score-reward mapping, and separately for regression and classification problems.

Finally, the historical regret used to define weights must incorporate a degree of historical smoothing, so that it accumulates a form of model reputation. This is accomplished using an exponential moving average (EMA) with a fiducial parameter $\alpha\in(0,1]$:
\be
\label{eq:net_regret}
{\cal R}_{il} = \alpha\left(\log{{\cal L}_{i}}-\log{{\cal L}_{il}}\right)+(1-\alpha){\cal R}_{i-1,l} ,
\ee
where the term $(\log{{\cal L}_{i}}-\log{{\cal L}_{il}})$ represents the current regret, i.e.\ the log-loss difference between the network loss and the individual worker loss (either for a raw or forecast-implied inference). The parameter $\alpha$ controls how strongly the network weighs historical performance. This too must be optimized independently for regression and classification problems.

\subsection{Summary of Incentive Structure and Reward Payouts} \label{sec:incentive}
In addition to the Inference Synthesis described above for regression and classification, the decentralized machine intelligence design presented by \citet{kruijssen24} rewards the participants according to their contributions to the network. We provide a short summary here and refer to \S4 and \S5 of \citet{kruijssen24} for further details.

Each individual worker is scored by quantifying its unique impact on the network loss (i.e.\ the log-loss difference incurred by removing it from the network), whereas reputers are scored by the proximity of their reported losses to the consensus. Rewards are distributed among workers by mapping their scores to a reward fraction, which is obtained by normalizing their scores by the standard deviation, applying the potential function of \autoref{eq:potential} (with the free parameter $p_{\rm i}$ and $p_{\rm f}$ setting the slope of the mapping for inferences and forecasts, respectively), and normalizing the result to unity so that we obtain reward fractions. Similarly, rewards are distributed among reputers by multiplying their scores with their stakes, raising these to a power $p_{\rm r}$, and again normalizing the result to unity so that we obtain reward fractions.

With the above definition of the reward distribution among similar network participants in place, the division of rewards between different network tasks (i.e.\ providing inferences, forecasts, and reputed losses) is performed based on a modified entropy of each task's reward fractions. The entropy quantifies the degree of decentralization by increasing with the number of participants and with the degree of similarity of their reward fractions. Tasks that are decentralized better get to distribute a larger share of the topic's rewards among their participants. The slopes $\{p_{\rm i}, p_{\rm f}, p_{\rm r}\}$ that characterize the aforementioned score-reward mapping (for inferences, forecasts, and reputed losses, respectively) affect the distribution of reward fractions, with higher slopes decreasing the entropy. Therefore, they do not only affect the reward fractions of the individual participants within a network task, but through the resulting entropy of these reward fractions also impact the distribution of rewards between the tasks.

Finally, the rewards are distributed between topics based on the topic weight, which is defined as the geometric average of the topic's total reputer stake and the recent fee revenue collected by a topic. This ensures that profitable topics backed by sufficient capital receive the highest rewards. It also generates a competition game, where it is in the interest of the topic participants to attract consumers are provide stake to the topic.

\section{Synthetic Data Generation of Ground Truth and for Workers and Reputers} \label{sec:ics}
The optimization of the network parameters is studied using a network simulator that implements the equations described in \citet{kruijssen24} for regression tasks, and \S\ref{sec:method} for classification tasks. The experiments are carried out using synthetic data sets that mimic (1) a ground truth time series and (2) the corresponding inferences provided by a set of workers participating in the network. The generation of these data sets differs considerably between simulations of regression and classification problems. In this section, we describe for each problem type how these data sets are generated.

\subsection{Regression} \label{sec:ics_regr}
The generation of mock data for regression tasks in our numerical experiments is summarized in \S3.3 of \citet{kruijssen24}, and we provide more details here. The experiments are carried out using a ground truth time series that mimics the price evolution of an asset $P_i$. We use the returns $\rho_i$ as the target variable for which the workers provide inferences, where $P_i$ and $\rho_i$ are related as
\be
\label{eq:price}
P_i = P_0\exp{\left(\sum_{i'\leq i}\rho_{i'}\right)} ,
\ee
and equivalently
\be
\label{eq:returns}
\rho_i = \ln{\left(\frac{P_i}{P_{i-1}}\right)} .
\ee
The returns $\rho_i$ are drawn at random from a normal distribution with mean $\mu_\rho=0.01$ and standard deviation $\sigma_\rho=0.01$, i.e.\ from ${\cal N}(0.01, 0.1)$, for a total duration of $N_{\rm epochs}=1000$ epochs.

The raw inferences produced by workers are generated as perturbations of the ground truth. Each individual inference-producing worker $j$ has an associated error $\sigma_j$, which is randomly drawn as
\be
\label{eq:sigmaj}
\log{\sigma_j} \leftarrow {\cal N}(\log{2\sigma_\rho}, \log{1.5}) = {\cal N}(-0.699, 0.176) ,
\ee
and an associated bias $\mu_j$, which is randomly drawn as
\be
\label{eq:muj}
\mu_j \leftarrow {\cal N}(0, 0.5\sigma_\rho) = {\cal N}(0,0.05) .
\ee
We assume that the worker gains experience and improves over time. This is modelled by multiplying its error and bias by a time-dependent factor $f_{{\rm xp},i}<1$:
\be
\label{eq:fxp}
f_{{\rm xp},i} = \frac{1}{2}\left(1+{\rm e}^{-0.03i}\right) ,
\ee
such that the error and bias experience an exponential decay from their intial value at $i=0$ to half that value at $i\gg30$. Additionally, at each epoch a single worker is selected at random and their error and bias are decreased by a multiplication with a factor of $f_{{\rm out},ij}=0.3$ to model a form of context-dependent outperformance, reflecting conditions under which that particular worker excels. All other workers have $f_{{\rm out},ij}=1$.

With the above setup, the simulation defines the raw inferences of \autoref{eq:inf_inference} as
\be
\label{eq:inf_inference2}
I_{ij} = \rho_i+\delta_{ij} ,
\ee
where the perturbation $\delta_{ij}$ is randomly generated for each worker $j$ at each epoch $i$ as
\be
\label{eq:deltaj}
\delta_{ij} \leftarrow {\cal N}(f_{{\rm out},ij}f_{{\rm xp},i}\mu_j,f_{{\rm out},ij}f_{{\rm xp},i}\sigma_j) .
\ee

The second task to which workers can contribute is to forecast the losses of the raw inferences (see \autoref{eq:fore_inference}). This follows a similar approach as for the inference task, in that each individual forecast-providing worker $k$ has an associated error $\sigma_k$, which is randomly drawn as
\be
\label{eq:sigmak}
\log{\sigma_k} \leftarrow {\cal N}(\log{0.6}, \log{1.5}) = {\cal N}(-0.222, 0.176) ,
\ee
and an associated bias $\mu_k$, which is randomly drawn as
\be
\label{eq:muk}
\mu_k \leftarrow {\cal N}(0, 0.3) .
\ee
Additionally, each forecasting worker has a context sensitivity parameter, which is obtained by randomly drawing a number from a uniform distribution $x_k\leftarrow{\cal U}(0,1)$ and applying a sigmoid function to create a bimodal distribution with increased incidence at the extremes of the interval $[0,1]$:
\be
\label{eq:fcontext}
f_{{\rm context},k} = \sigma[10(x_k-0.5)] .
\ee
With this setup, the simulation defines the forecasted losses from \autoref{eq:fore_inference} as
\be
\label{eq:fore_inference2}
\log{L_{ijk}} = {\cal L}_{ijk} + \delta_{ik} ,
\ee
where $\delta_{ik}$ is generated analogously to \autoref{eq:deltaj}:
\be
\label{eq:deltak}
\delta_{ik} \leftarrow {\cal N}(f_{{\rm xp},i}\mu_k,f_{{\rm xp},i}\sigma_k) ,
\ee
and the target loss ${\cal L}_{ijk}$ of forecaster $k$ is obtained through interpolation between the true loss including and excluding contextual outperformance. The first is obtained by evaluating the loss function ${\cal L}$ by using the true inference $I_{ij}$ from \autoref{eq:inf_inference2}, whereas the second is obtained by instead using a modified inference $I'_{ij}$ that would have been generated had we set $f_{{\rm out},ij}=1~\forall~\{i,j\}$ in \autoref{eq:deltaj}:
\be
\label{eq:trueloss}
\log{\cal L}_{ijk} = f_{{\rm context},k}\log{\cal L}(I_{ij}) + (1-f_{{\rm context},k})\log{\cal L}(I'_{ij}) .
\ee
This way, the forecasted losses are a perturbation of the actual loss, modulated by the context awareness of the worker as to whether or not the inference outperformed. This is a simple way of modelling the forecast-providing workers' attempts to predict the quality of raw inferences under varying contextual circumstances.

Finally, reputers in the simulation are initialized similarly as before, i.e.\ by associating each reputer $m$ with an error $\sigma_m$, which is randomly drawn as
\be
\label{eq:sigmam}
\log{\sigma_m} \leftarrow {\cal N}(\log{0.1}, \log{1.25}) = {\cal N}(-1.000, 0.097) ,
\ee
and a bias $\mu_m$, which is randomly drawn as
\be
\label{eq:mum}
\mu_m \leftarrow {\cal N}(0, 0.05) .
\ee
Additionally, each reputer is associated with an evolving stake $S_{im}$, of which the initial stake $S_{0m}$ is randomly generated from a Pareto distribution ${\cal P}(2,1.6\times10^5)$ with slope $\alpha=2$ and minimum stake $1.6\times10^5$~ALLO.

With the above setup, the vector of losses $\vec{L}_{im}$ reported by reputer $m$ at epoch $i$ follows as
\be
\label{eq:rep_loss}
\log{\vec{L}_{im}} = \log{\vec{L}_{i}} + \delta_{im} ,
\ee
where $\vec{L}_{i}$ is the true loss vector and the perturbation $\delta_{im}$ is randomly generated for each reputer as
\be
\label{eq:deltam}
\delta_{im} \leftarrow {\cal N}(\mu_m,\sigma_m) .
\ee
Contrary to the discussion in \S3.3 of \citet{kruijssen24}, we do not let reputers apply an EMA to their reported losses (equivalent to setting $\alpha_m=1$ there), because the forecast-providing workers require unaltered losses to maximize their context awareness.

\subsection{Classification} \label{sec:ics_class}
The mock data generation process for classification tasks is highly similar to the procedure described in \S\ref{sec:ics_regr}, with the exception of the generation of the ground truth, which naturally differs between regression and classification tasks. For classification tasks, the ground truth represents a unit vector with length equal to the number of labels $N_{ic}$ at epoch $i$. In other words, once a label occurs it has a probability of unity and all other labels have a probability of zero. As discussed in \S\ref{sec:method_class}, models provide classification inferences by assigning probabilities to each label $c$ on the domain $P_c\in[0,1]$. To enable the straightforward generation of mock inferences, we choose to not generate the ground truth as a unit vector, but as a set of real-valued probabilities, of which the largest value is considered to occur in reality. The raw inferences can then be generated as perturbations of the ground truth probability vector, analogously to the procedure described in \S\ref{sec:ics_regr} (see \autoref{eq:inf_inference2} in particular).

\begin{figure}
    \centering
    \includegraphics[width=0.375\linewidth]{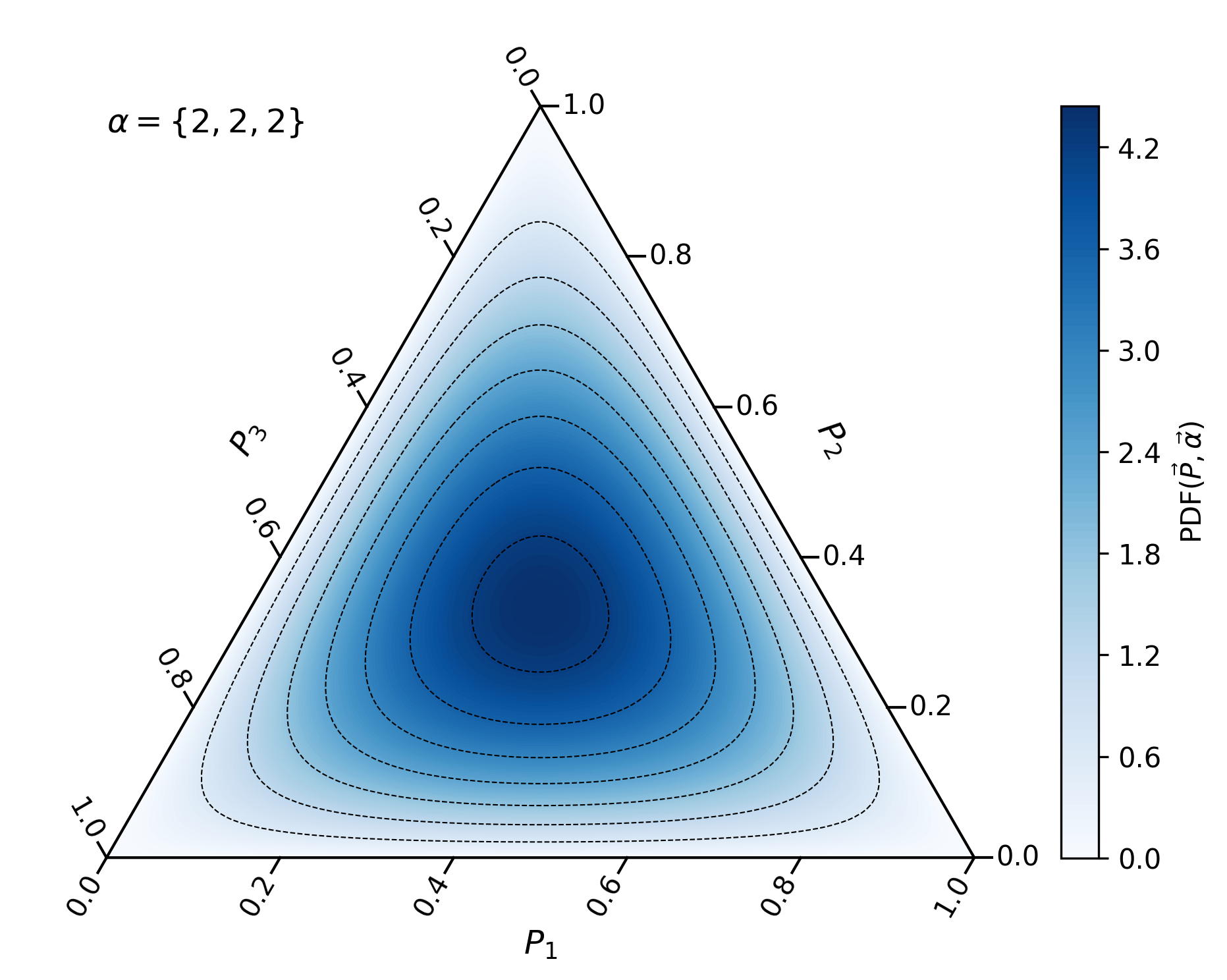}%
    \hspace{-0.03\textwidth}
    \includegraphics[width=0.375\linewidth]{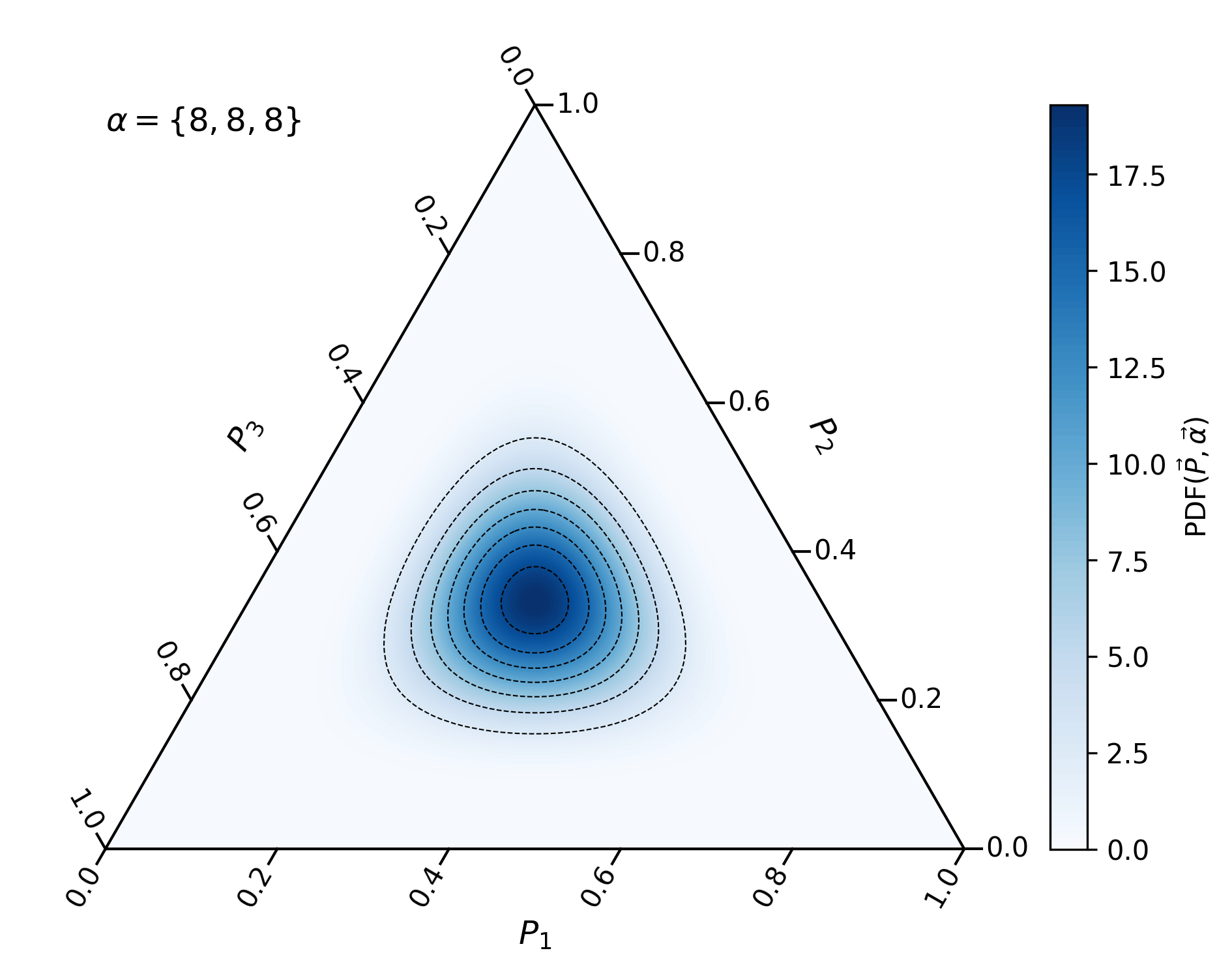}%
    
    \includegraphics[width=0.375\linewidth]{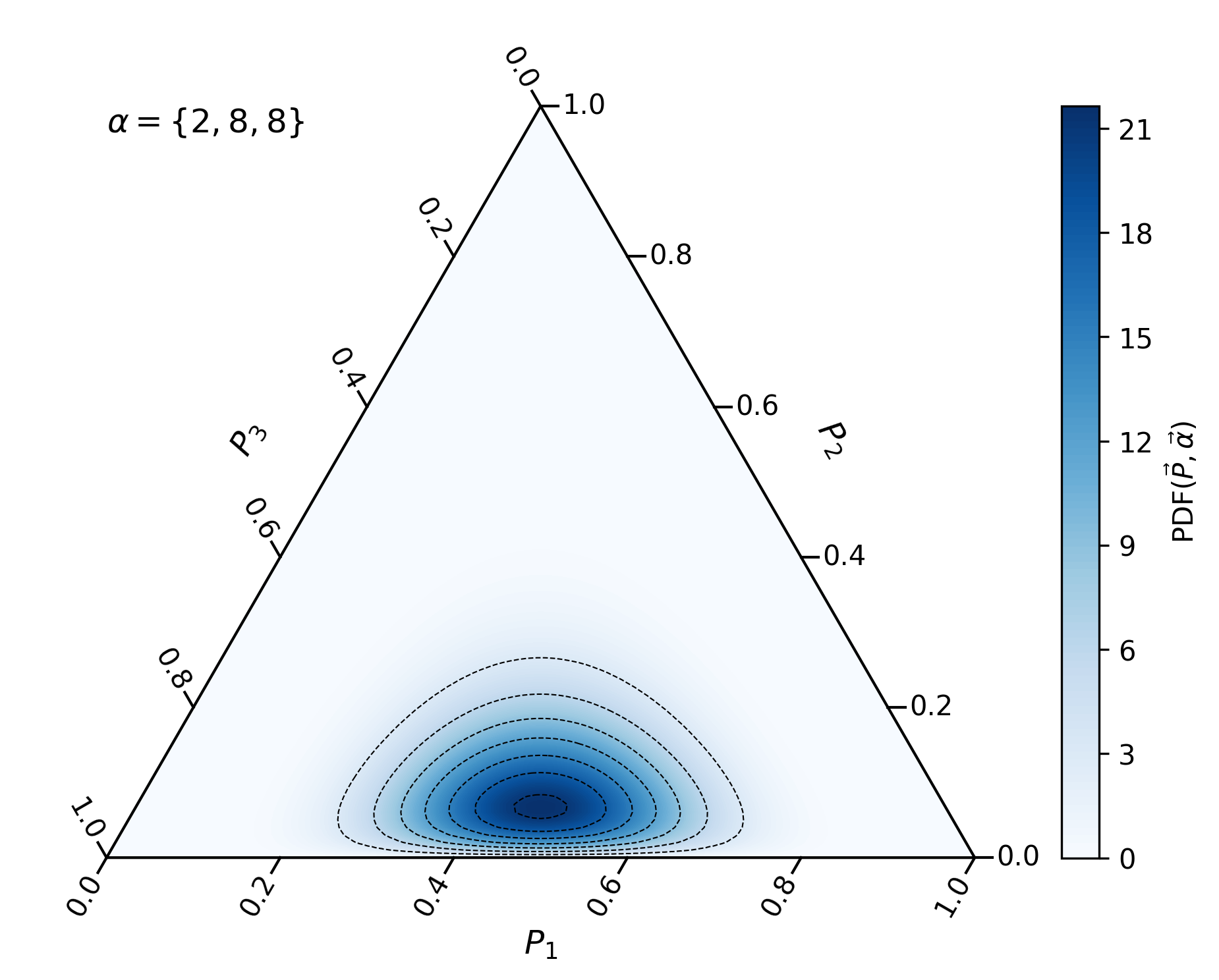}%
    \hspace{-0.03\textwidth}
    \includegraphics[width=0.375\linewidth]{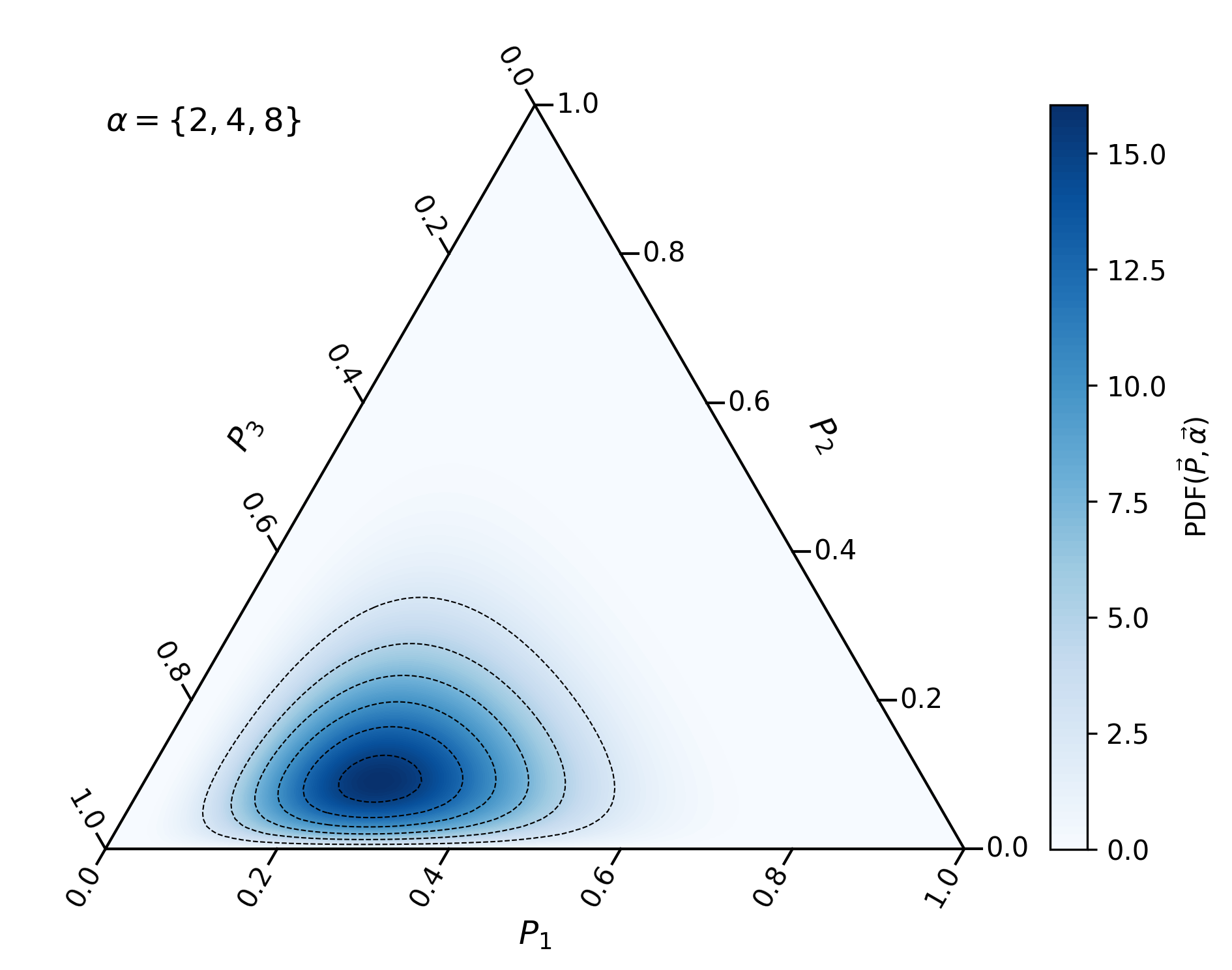}%
    \caption{Probability distributions generated from the Dirichlet distribution ${\cal D}$ for a set of three labels $N_{ic}=3$ (where $\sum_c P_{ic}=1$ implies that the parameter space is a two-dimensional triangle), using different concentration parameter vectors $\vec{\alpha}$. A comparison between the two top panels shows that the normalization of $\vec{\alpha}$ controls the overall concentration of probabilities, with higher $\norm{\vec{\alpha}}/\sqrt{N_{ic}}$ resulting in a higher degree of concentration. The two bottom panels illustrate how the probabilities become unbalanced if one (bottom-left) or all three (bottom-right) concentration parameters differ.
    }
    \label{fig:dirichlet}
\end{figure}
The ground truth is a probability vector with length $N_{ic}$, denoted as $\vec{P}_i$. The probability vector is drawn from a Dirichlet distribution ${\cal D}$ \citep[e.g.][]{olkin64}, which is a multivariate generalization of the beta distribution and is parameterized with a vector $\vec{\alpha}$ that encodes the concentration of each label. We illustrate the behavior of the Dirichlet distribution in \autoref{fig:dirichlet}, showing that setting $\vec{\alpha}$ enables full control over the (im)balance of the labels as well as the difficulty (probability concentration) of the inference problem. For simplicity, we use a two-parameter definition of $\vec{\alpha}$, by modifying an index vector of length $N_{ic}$ with a variation parameter $v\in[0,1]$ (controlling the overall concentration) and a balance parameter $b\in[0,1]$ (controlling the balance of the label probabilities), resulting in:
\be
\label{eq:vecalpha}
\vec{\alpha}_i = \left(\frac{2\{1, 2, \ldots, N_{ic}\}}{N_{ic}+1}\right)^{2(1-\sqrt{b})}\times10^{2(1-2v)} .
\ee
This specific parameterization is far from unique and other forms are certainly possible. The division of the index vector by $(N_{ic}+1)/2$ is equivalent to a normalization to the median, ensuring it is symmetric around unity. The dependence of the exponent on $b$ lets $b$ control the amplitude of the deviations from unity, and the multiplication with a $v$-dependent term lets $v$ control the normalization. The numerical factors are chosen to enable a clean mapping from the parameter domain $[0,1]$. Comparison with \autoref{fig:dirichlet} shows that this form achieves the desired behavior that high $v$ results in a low normalization of $\vec{\alpha}$ and thus a high degree of variation, as well as that high $b$ results in a small dynamic range of $\vec{\alpha}$ and thus a high balance. With this setup, we arrive at the generation of the ground truth probability vector:
\be
\label{eq:pvec}
\vec{P}_i \leftarrow {\cal D}[\vec{\alpha}_i(b,v)] ,
\ee
which is reduced to a unit vector expressing the occurrence of the actual ground truth as
\be
\label{eq:pvec2}
\vec{\cal P}_i = \{{\cal P}_1, {\cal P}_2, \dots, {\cal P}_{N_{ic}}\}, \quad \text{where} \quad {\cal P}_i = 
\begin{cases} 
1 & \text{if } i = \arg\max_{i'}\vec{P}_{i'}, \\
0 & \text{otherwise}.
\end{cases}
\ee

Simulated network participants are initialized in the same way as for regression tasks (see \autoref{eq:sigmaj}--\autoref{eq:fxp} for inference-producing workers, \autoref{eq:sigmak}--\autoref{eq:fcontext} for forecast-producing workers, and \autoref{eq:sigmam}-\autoref{eq:mum} for reputers). The raw inferences are generated as
\be
\label{eq:inf_inference_class2}
\vec{I}_{ij} = \frac{\sigma{\left\{\lambda{(\vec{P}_i)} + \left[\frac{1}{\vec{P}_i\circ(1-\vec{P}_i)}\right]\circ\vec{\delta}_{ij}\right\}}}{\sum_c\sigma{\left\{\lambda{(\vec{P}_i)} + \left[\frac{1}{\vec{P}_i\circ(1-\vec{P}_i)}\right]\circ\vec{\delta}_{ij}\right\}}} ,
\ee
where $\sigma$ is the sigmoid function, $\lambda$ is its inverse (i.e.\ the logit function), the factor of $1/\vec{P}_i\circ(1-\vec{P}_i)$ accounts for the slope of the logit function, $\circ$ indicates element-wise multiplication, and $\sum_c$ indicates taking the sum over all vector elements, ensuring that $\sum_c\vec{I}_{ij}=1$. The sigmoid-logit transformations ensure that the perturbation by $\vec{\delta}_{ij}$ cannot result in probabilities outside of the allowed range ($P_{ic}\in[0,1]~\forall~c$). We generate the perturbation vector $\vec{\delta}_{ij}$ on an element-by-element basis as
\be
\label{eq:deltaj_class}
\vec{\delta}_{ij} \leftarrow {\cal N}(f_{\rm class}f_{{\rm out},ij}f_{{\rm xp},i}\mu_j,f_{\rm class}f_{{\rm out},ij}f_{{\rm xp},i}\sigma_j) ,
\ee
where the extra term $f_{\rm class}=3$ is a boost factor to achieve a degree of stochasticity that is similar between regression and classification tasks.

All other synthetic data generation steps match the description for regression from \S\ref{sec:ics_regr}. Together, they represent the initial conditions for the Inference Synthesis process described in \S\ref{sec:method}, as well as the incentive structure presented in \S4 of \citet{kruijssen24} and summarized in \S\ref{sec:incentive}.

\section{Optimizing Performance for Inference Synthesis and Reward Distribution} \label{sec:opt}
We now turn to the optimization of the free parameters that exist in the infrastructure for decentralized, online learning described in \S\ref{sec:method}. We determine the parameter sets that maximize network performance, and how these vary as a function of network composition. Note that the optimization performed here is not restricted to the specific experimental setup of this work, but is generalizable to any form of decentralized machine intelligence that performs Inference Synthesis through a form of performance-to-weight mapping and historical reputation, or provides a form of monetary incentive to its participants.

\subsection{Parameter Optimization} \label{sec:param_opt}
\begin{table}[b]
  \centering
  \begin{minipage}{0.48\textwidth}
    \centering
    \begin{tabular}{ccccc}
        \hline
        Parameter & Minimum & Maximum & Step Size & Default \\
        \hline
        $p$ & 1.0 & 5.0 & 0.5 & 3.0 \\
        $\log_{10}(\alpha)$ & $-$5.0 & 0.0 & 0.5 & $-$1.0 \\
        $p_{\rm i}$ & 1.0 & 5.0 & 1 & 3.0 \\
        $p_{\rm f}$ & 1.0 & 5.0 & 1 & 3.0 \\
        $p_{\rm r}$ & 0.5 & 1.5 & 0.25 & 1.0 \\
        \hline
    \end{tabular}
    \caption{Parameter sets considered.}
    \label{tab:param_opt}
  \end{minipage}%
  \hfill
  \begin{minipage}{0.48\textwidth}
    \centering
    \begin{tabular}{cccl}
        \hline
        $N_{\rm i}$ & $N_{\rm f}$ & $N_{\rm r}$ & Comment \\
        \hline
        5 & 3 & 5 & default \\
        10 & 5 & 5 &  \\
        15 & 7 & 5 &  \\
        20 & 10 & 5 &  \\
        25 & 12 & 5 &  \\
        30 & 15 & 5 & high numbers of workers \\
        \hline
    \end{tabular}
    \caption{Network compositions considered.}
    \label{tab:net_comp}
  \end{minipage}
\end{table}
The main free parameters that we optimize are the slopes of the potential function used for regret-to-weight mapping ($p$) and score-to-reward mapping ($p_{\rm i}$, $p_{\rm f}$, and $p_{\rm r}$ for inferences, forecasts, and reputations, respectively), as well as the exponential moving average parameter $\alpha$ that weights the historical performance of the network. Increasing $p$ results in a steeper mapping of regrets to weights, effectively turning the Inference Synthesis process increasingly into a model selection process. Increasing $p_{\rm i}$, $p_{\rm f}$, and $p_{\rm r}$ results in a steeper mapping of scores to rewards, effectively turning the reward distribution process into a more aggressive winner-takes-all process. Increasing $\alpha$ results in a stronger weighting of recent performance in the regret (and thus weight) calculation, resulting in more stochasticity, but also in more adaptive behavior.

We perform the optimization for both regression and classification tasks, and for different network compositions (in \S\ref{sec:net_comp} below). The objective function that we optimize is the network loss, defined for regression as the mean squared error between the network inference and the returns:
\be
\label{eq:net_loss} 
{\cal L}_i = \left(I_i-\rho_i\right)^2 ,
\ee
where $I_i$ is the network inference and $\rho_i$ is the returns at epoch $i$ (see \autoref{eq:returns}). Similarly, for classification tasks, we optimize a modified version of the mean squared error between the network inference and the ground truth:
\be
\label{eq:net_loss_class}
{\cal L}_i = \sum_c\left(\vec{I}_{i}-\vec{\cal P}_{i}\right)^2 ,
\ee
where $\sum_c$ indicates taking the sum over all vector elements, ensuring that $\sum_c\vec{I}_{i}=1$, and $\vec{\cal P}_i$ is the ground truth probability vector at epoch $i$ (see \autoref{eq:pvec2}). The advantage of using the mean squared error even for classification tasks is that it accounts also for the inferred probabilities of the incorrect labels. It also allows us to compare the performance of the two different types of tasks using the same metric.

For both regression and classification, we optimize the parameters by iterating over a simple parameter grid, and for each parameter set we compute a vector of network losses over 1000 epochs. The parameter sets that are evaluated are listed in \autoref{tab:param_opt}. Each parameter set is evaluated for 11 different network compositions, which are listed in \autoref{tab:net_comp}. Across all compositions, we first focus on the variation in network loss as a function of $p$ and $\log_{10}(\alpha)$.

\begin{figure}
  \centering
  \includegraphics[width=0.825\linewidth]{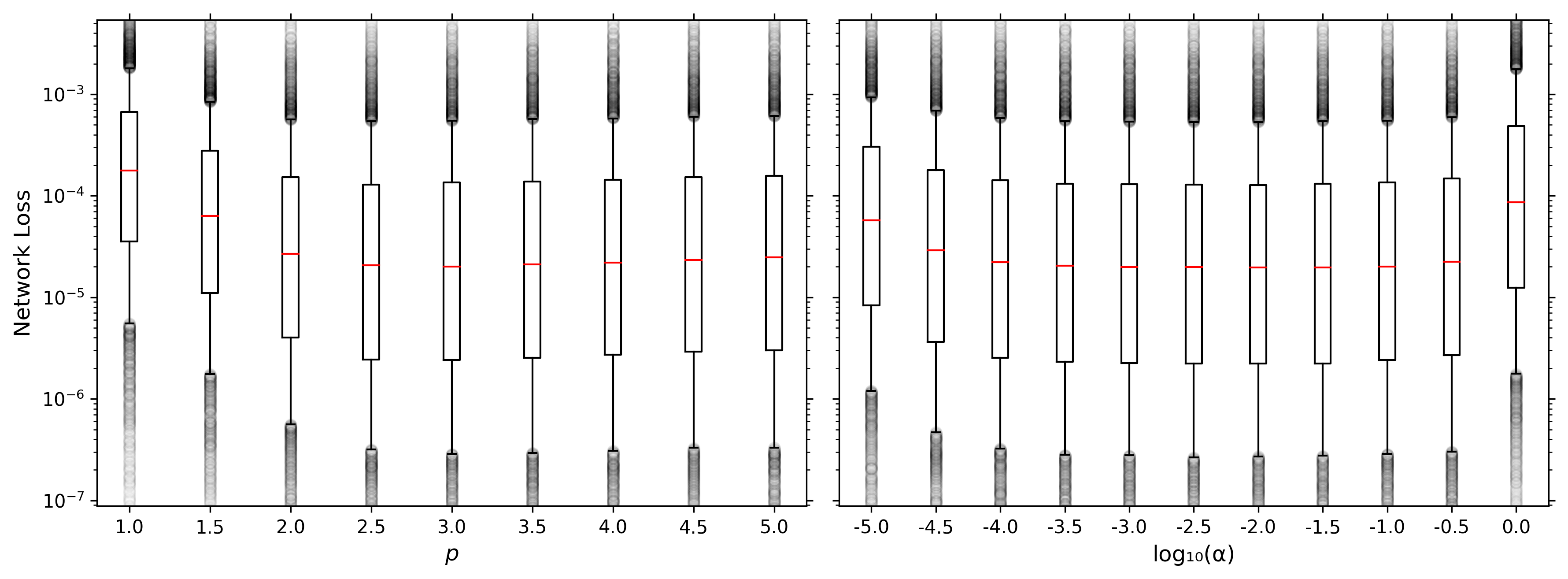}%
  
  \includegraphics[width=0.825\linewidth]{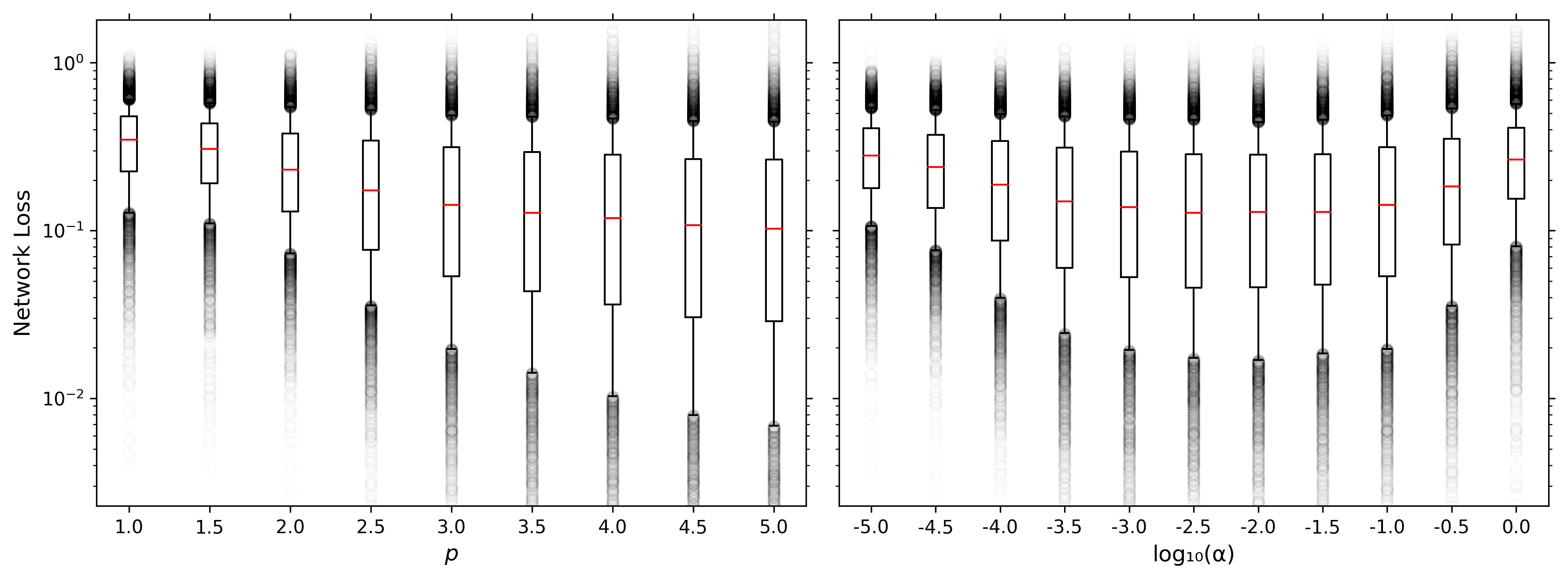}%
  \caption{Variation in network loss as a function of the regret-to-weight mapping slope $p$ (left) and the EMA parameter $\log_{10}(\alpha)$ (right), for regression (top) and classification (bottom). The box plots show the network loss over 1000 epochs and across all network compositions, with a red line showing the median and whiskers extending to the 10th and 90th percentiles. Beyond these percentiles, outliers are shown as transparent individual points. Based on these experiments, we select default parameters $p=3$ for regression, $p=5$ for classification, and $\log_{10}(\alpha)=-1.0$ (see the text for discussion).
  }
  \label{fig:network_loss}
\end{figure}

\autoref{fig:network_loss} shows how the network loss depends on $p$ and $\log_{10}(\alpha)$ for regression and classification tasks. Across the full parameter space considered here, we identify considerable scatter. Focusing on the median network loss as a function of $p$ and $\log_{10}(\alpha)$ to minimize the effect of outliers, we find that the optimal values of $p$ and $\alpha$ differ considerably between regression and classification tasks. For regression, the network loss is minimized for $p=3$, which was the default adopted by \citet{kruijssen24}. For classification, the network loss monotonically decreases with increasing $p$. Although this is not shown in \autoref{fig:network_loss}, we find that this decrease saturates for $p>5$. We will show in \S\ref{sec:opt} that it is not beneficial to increase $p$ beyond $p=5$ for classification tasks. The reason for the difference between regression and classification tasks is plausibly that the dynamic range of the losses is smaller for classification, such that the regret-weight mapping requires more sensitive differential weighting.

For both regression and classification, the network loss depends on $\alpha$ in a similar fashion, with a broad trough between $\log_{10}(\alpha)=-3.5$ and $\log_{10}(\alpha)=-1.0$, being somewhat wider for regression. A lower $\alpha$ results in a stronger weighting of historical performance, which is beneficial for the network's memory, but results in a less responsive network that is slower to adapt to changes in performance. The results thus illustrate that values of $\alpha$ so low that they cannot impact the network within the 1000-epoch duration of the experiments result in elevated network loss. Conversely, a higher $\alpha$ results in a weaker weighting of historical performance, which is beneficial for the network's adaptability, but in the extreme case of $\alpha=1$ results in a complete disregard for historical performance and a correspondingly higher network loss. It is the interest of the network's adaptability to maximize $\alpha$ while minimizing the network loss. In view of the wide trough in the network loss observed here, we identify $\log_{10}(\alpha)=-1.0$ (or $\alpha=0.1$) as a good balance between adaptability and historical awareness.

In order to assess the score-to-reward mapping slopes $p_{\rm i}$, $p_{\rm f}$, and $p_{\rm r}$, we aim to minimize the spread in the mean reward received per participant across the three classes of activity. The reason for this minimization is that given the same performance (as is the case in our experiments), participants are expected to receive the same reward independently of the activity they perform. As such, the observed spread in reward is solely determined by the score-to-reward mapping parameters. We quantify the spread in reward as the instantaneous standard deviation of the mean reward received across each class of activity at epoch $i$, i.e.
\be
\label{eq:std_reward}
\sigma_i = \sigma_{\rm tasks}\left\{\frac{1}{N_{\rm i}}\sum_j u_{ij}, \frac{1}{N_{\rm f}}\sum_k v_{ik}, \frac{1}{N_{\rm r}}\sum_m w_{im}\right\} ,
\ee
where $\sigma_{\rm tasks}$ indicates taking the standard deviation over the three tasks, the curly brackets collect the tasks in a vector, and the quantities $u_{ij}$, $v_{ik}$, and $w_{im}$ represent the reward received at epoch $i$ by individual inferers, forecasters, and reputers, respectively.

\begin{figure}
    \centering
    \includegraphics[width=0.9\linewidth]{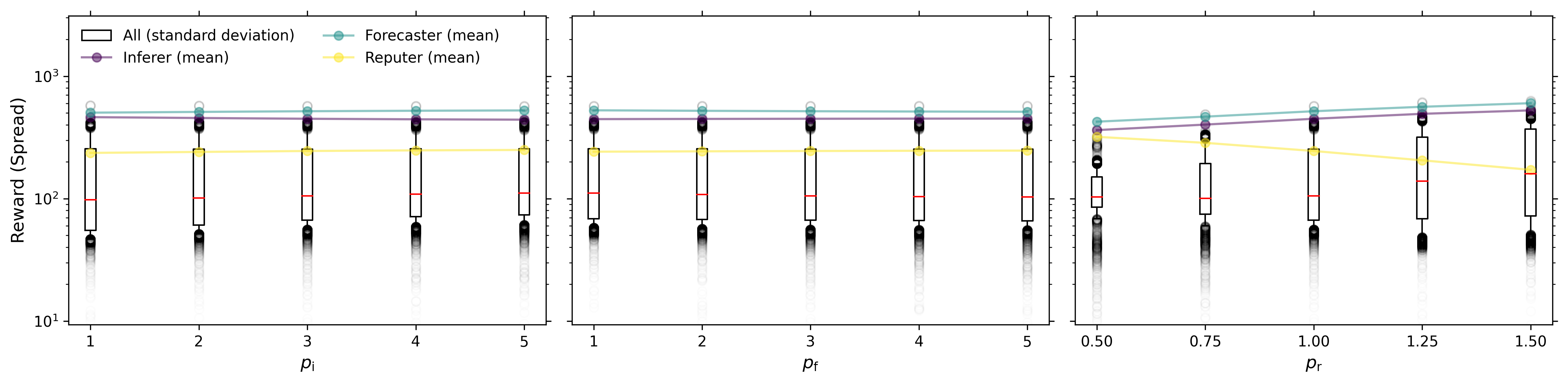}%
    
    \includegraphics[width=0.9\linewidth]{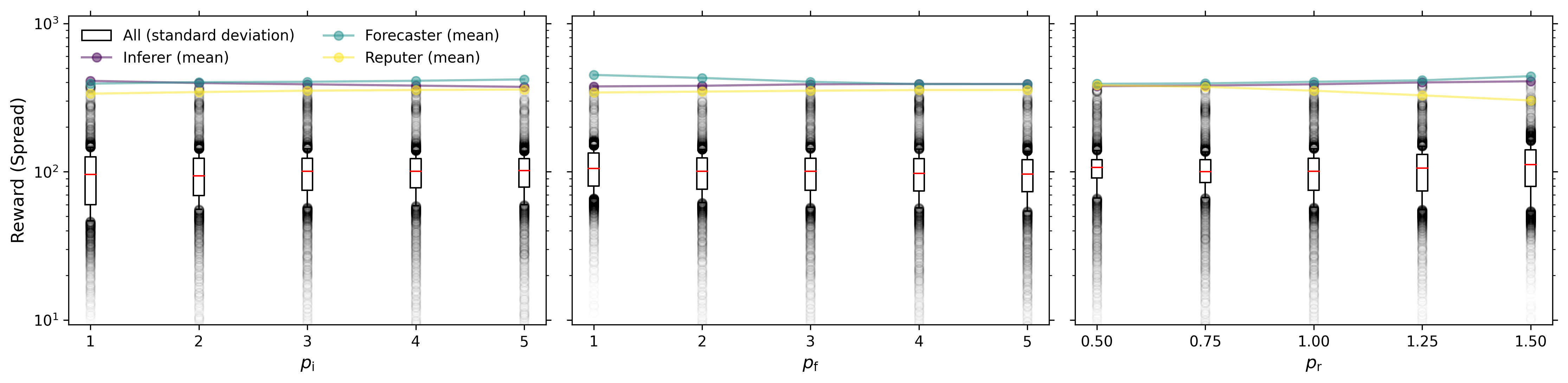}%
    \caption{Variation in the instantaneous standard deviation of the mean reward received across each class of activity at epoch $i$, as a function of the score-to-reward mapping slope $p_{\rm i}$ (left), $p_{\rm f}$ (middle), and $p_{\rm r}$ (right), for regression (top) and classification (bottom). The box plots show the reward spread over 1000 epochs and across all network compositions, with a red line showing the median and whiskers extending to the 10th and 90th percentiles. Beyond these percentiles, outliers are shown as transparent individual points. For comparison, the mean reward received by a participant in each class is shown as a colored line, with the color reflecting the activity type as indicated by the legend. Based on these experiments, we select the default parameters $p_{\rm i}=3$, $p_{\rm f}=3$, and $p_{\rm r}=1$ (see the text for a discussion).
    }
    \label{fig:reward_spread}
\end{figure}
The dependence of the reward spread on the score-to-reward mapping slopes $p_{\rm i}$, $p_{\rm f}$, and $p_{\rm r}$ is shown in \autoref{fig:reward_spread}. When steepening the score-to-reward mapping slope for any class of participant, we see that its mean reward decreases. This decrease is driven by the fact that the reward fraction allocated to each class increases with its entropy, aiming to incentivize decentralization. With steeper score-to-reward mapping slopes, the reward distribution within a class becomes less equal, causing the entropy to decrease and the reward allocation of that class to decrease. Interestingly, the impact of any of the two worker slopes ($p_{\rm f}$ and $p_{\rm r}$) on the standard deviation of the mean reward is minor ($<12\%$ for regression, $<8\%$ for classification) across the considered range of $p_{\rm i}$ and $p_{\rm f}$, as each of the three means are nearly constant over this interval. Because the experiments have a higher number of inferers than forecasters, the mean reward per participant is higher for forecasters. The entropy dependence of reward distribution thus causes the mean rewards per participant of inferers and forecasters to become more equal at low $p_{\rm i}$ or high $p_{\rm f}$. These opposite trends in mean reward favor adopting similar slopes for inferers and forecasters.

The results for reputers differ, as the reward calculation differs from those of inferers and forecasters (see \S\ref{sec:incentive}) and we find a steeper decline of the mean reward with increasing $p_{\rm r}$. The spread of the mean reward across all three classes of activity is determined by the interplay between differing absolute numbers of participants, the score-to-reward mapping slopes, and the corresponding reward entropies. For the experiments performed here, we find that the reward spread is minimized for $p_{\rm r}=1$, but even when varying $p_{\rm r}$ over a wide range of values, the dynamic range of the reward spread is moderate ($<45\%$ for regression) to small ($<12\%$ for classification).

The above results provide a good basis for the selection of the default parameters $p=3$ (regression) or $p=5$ (classification), $\log_{10}(\alpha)=-1.0$, $p_{\rm i}=3$, $p_{\rm f}=3$, and $p_{\rm r}=1$, across a broad range of network compositions. While some trends exist between reward spread and the score-to-reward mapping slope of reputers, the dynamic range of the reward spread remains sufficiently small to favor simplicity over optimization.

\subsection{Network Composition} \label{sec:net_comp}
The next key question is how the network loss and reward spread depend on the network composition, i.e.\ the number of participants in each class of activity. We again consider the network compositions listed in \autoref{tab:net_comp}, and for each composition we determine the network loss and reward spread as a function of the parameter sets listed in \autoref{tab:param_opt}. As discussed in the context of \autoref{fig:reward_spread}, the network composition determines the relative numbers of participants in each class of activity, and thus the entropy of the reward distribution. To isolate this effect, we adjust the number of inferers and forecasters in conjunction, while keeping the number of reputers fixed.

\begin{figure}
  \centering
  \includegraphics[width=0.9\linewidth]{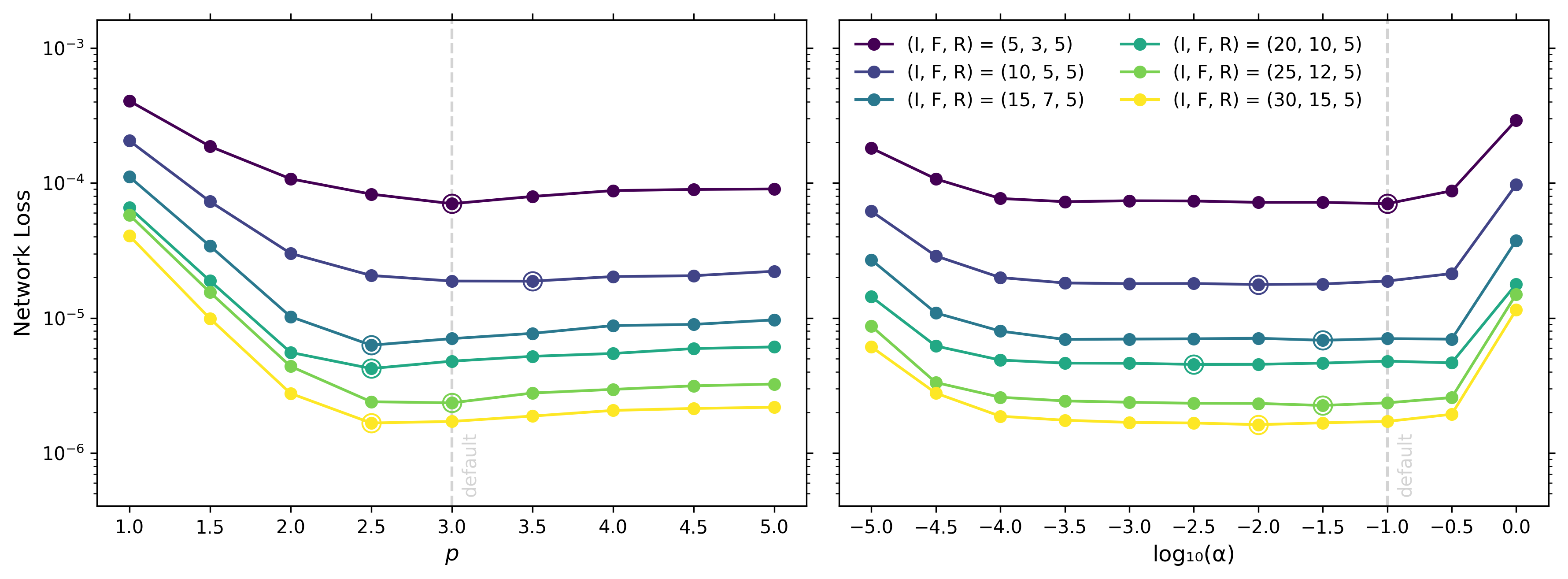}%
  
  \includegraphics[width=0.9\linewidth]{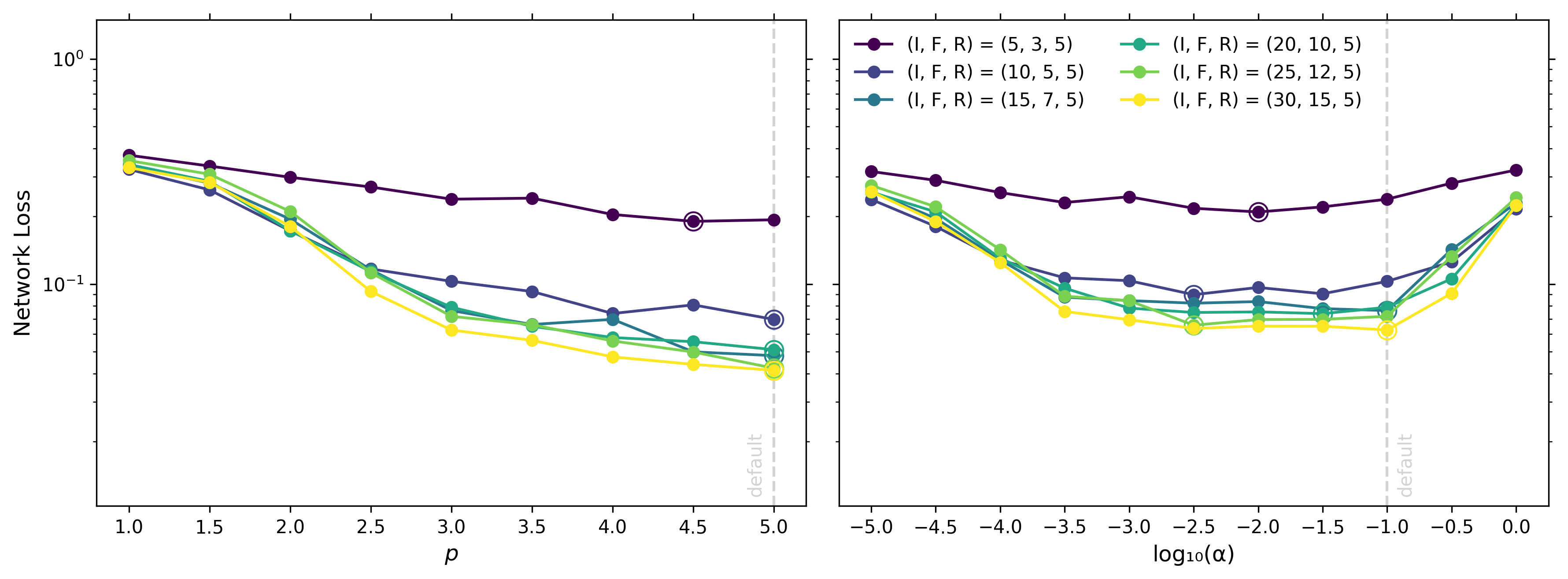}%
  \caption{Variation in median network loss for different network compositions, as a function of the regret-to-weight mapping slope $p$ (left) and the EMA parameter $\log_{10}(\alpha)$ (right), for regression (top) and classification (bottom). The median is taken over 1000 epochs, with each line representing a different network composition as indicated by the legend. For each network composition, an open circle marks the $x$-value of the minimum network loss for that composition. The vertical dashed lines show the default values, i.e.\ $p=3$ for regression, $p=5$ for classification, and $\log_{10}(\alpha)=-1.0$.
  }
  \label{fig:network_loss_composition}
\end{figure}

\autoref{fig:network_loss_composition} shows the network loss as a function of the regret-to-weight mapping slope $p$ and the EMA parameter $\log_{10}(\alpha)$ for the different network compositions. The overall trends between the network loss and network parameters are similar to those shown in \autoref{fig:network_loss} when integrating over all network compositions. Interestingly, the clear trend of network loss decreasing with increasing numbers of participants in \autoref{fig:network_loss_composition} reveals that a considerable part of the scatter in network loss found in \autoref{fig:network_loss} can be attributed to differences in network composition. It is not surprising that the network loss is generally lower for higher numbers of inferers and forecasters, as a wider variety of inferences and forecasts are available to the network, and the inferred returns are thus more accurate. For classification tasks, the positive impact of increasing the number of participants on the network loss vanishes beyond $N_{\rm i}\sim10$ and $N_{\rm f}\sim5$, as adding more inferences and forecasts does not further improve the determination of the highest-probability label. This is a fundamental difference between regression and classification tasks, and it underlines the importance of the task type in determining the network's optimal composition. Classification tasks benefit from more stringent limits on the number of participants (e.g.\ $N_{\rm i}=10$ and $N_{\rm f}=5$) than regression tasks (e.g.\ $N_{\rm i}=30$ and $N_{\rm f}=15$).

The dependence of the minimum network loss on the network parameters $p$ and $\log_{10}(\alpha)$ is similar to that in \autoref{fig:network_loss} when considering different network compositions. The optimal regret-to-weight mapping slope is still $p=3$ for regression and $p=5$ for classification. Similarly, the optimal EMA parameter $\log_{10}(\alpha)$ is still $-1.0$ for both regression and classification, even if the interval over which it could potentially be varied is narrower for higher numbers of participants. This is a convenient result, as it allows for the use of a single set of default parameters, independently of the network composition.

\begin{figure}
    \centering
    \includegraphics[width=0.5\linewidth]{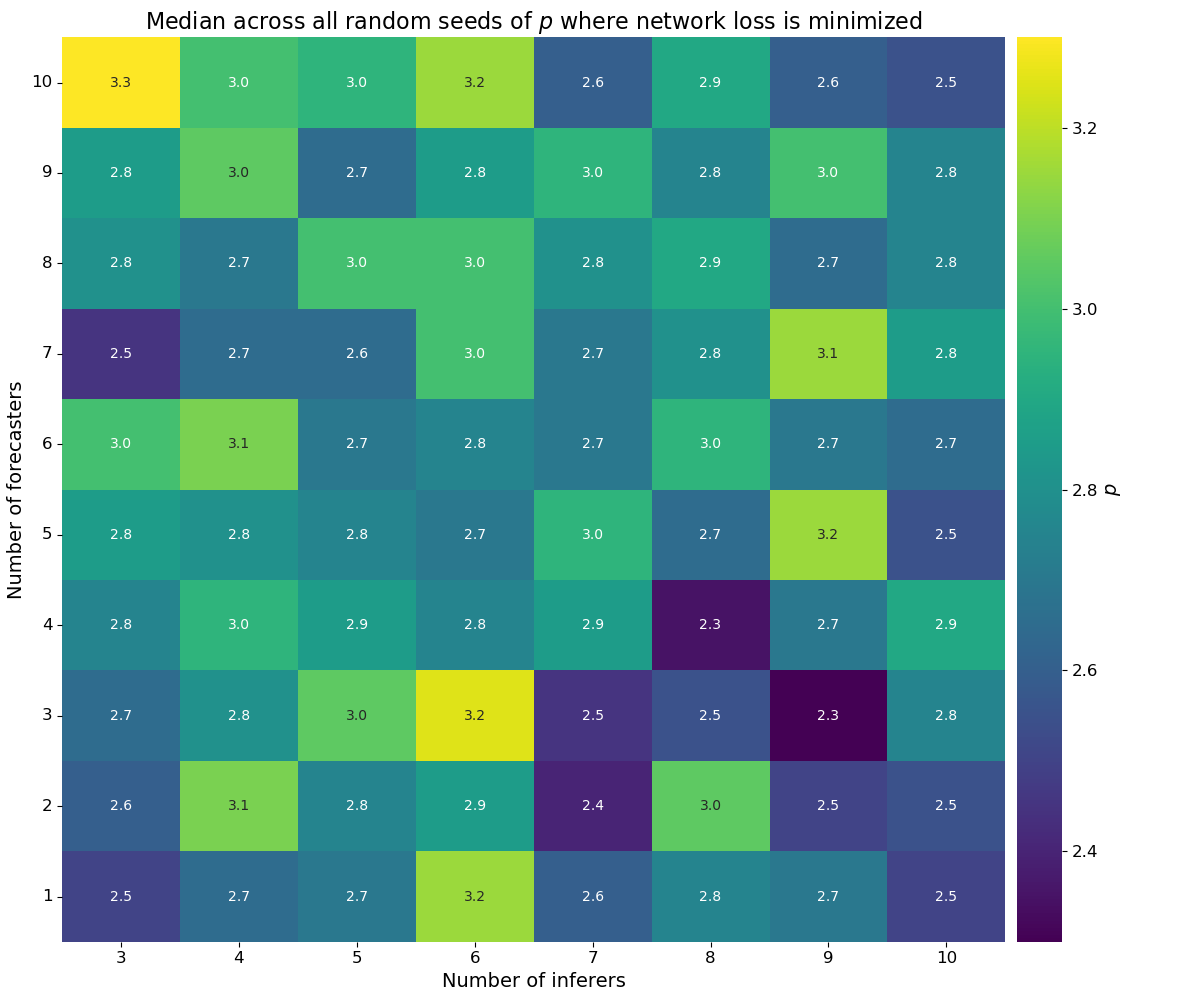}%
    \includegraphics[width=0.5\linewidth]{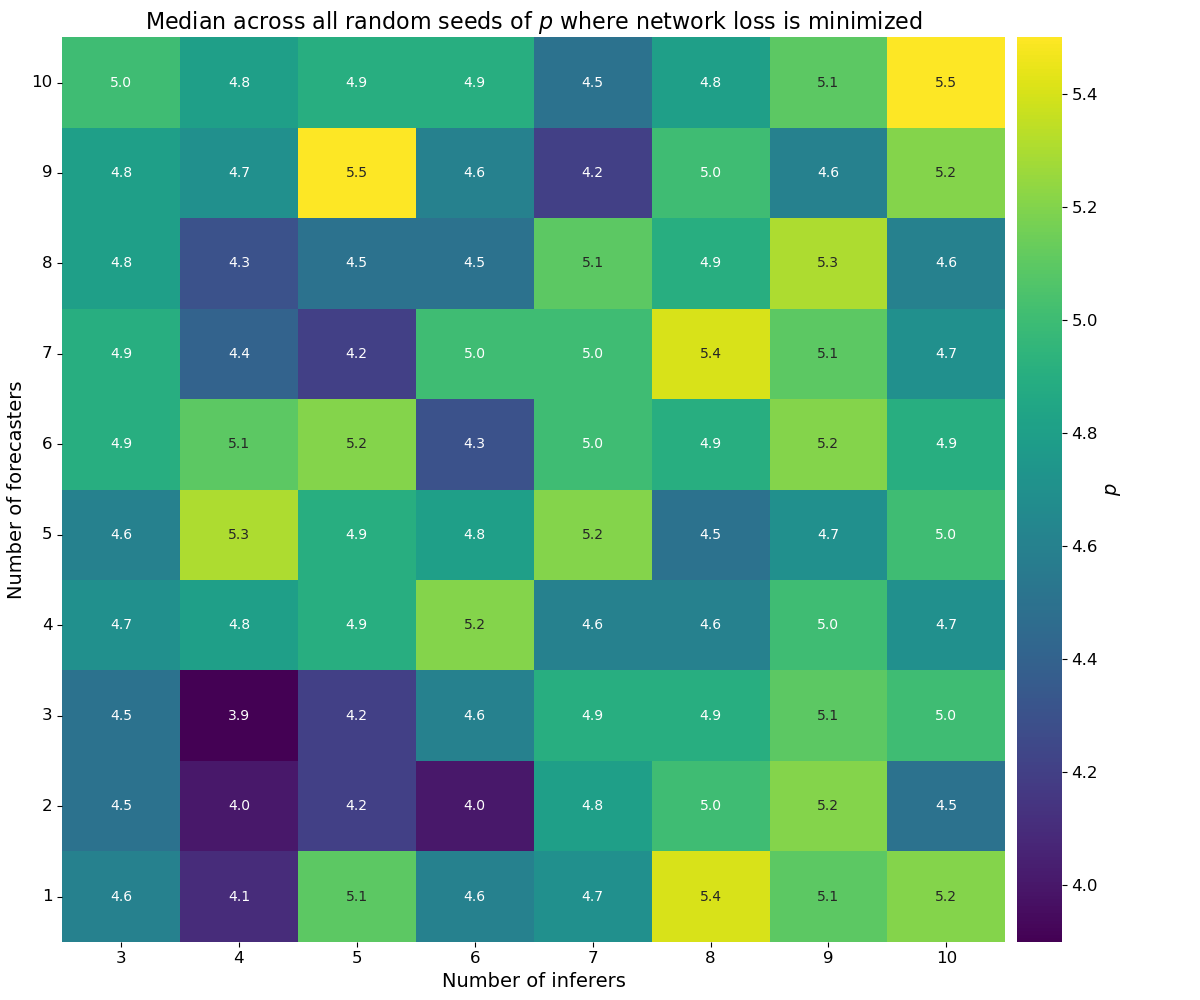}%
    \caption{Variation of the regret-to-weight mapping slope $p$ that minimizes the network loss with network composition, represented as a function of the number of inferers $N_{\rm i}$ and forecasters $N_{\rm f}$ for regression (left) and classification (right). Each value is obtained by taking the median over 10 experiments with different random seeds. There might be a hint of steeper regret-to-weight mappings for inferer-heavy network compositions (when $N_{\rm i}>N_{\rm f}$, we have mean $\overline{p}=2.81\pm0.03$ for regression and $\overline{p}=4.85\pm0.05$ for classification) than for forecaster-heavy network compositions (when $N_{\rm f}>N_{\rm i}$, we have mean $\overline{p}=2.79\pm0.04$ for regression and $\overline{p}=4.70\pm0.08$ for classification). The absence of any major trends between $p$ and network composition implies that the optimal regret-to-weight mapping slope $p$ can be set to the default values of $p=3$ (regression) or $p=5$ (classification), independently of the network composition.
    }
    \label{fig:heatmaps}
\end{figure}
To systematically assess the impact of the network composition on the regret-to-weight mapping slope $p$ that optimizes the network loss and to further refine and generalize our findings, we repeat the network loss minimization for 10 different random seeds across the complete grid spanned by $N_{\rm i}=3{-}10$ and $N_{\rm f}=1{-}10$, modifying $p$ from $2.0{-}3.6$ in steps of $0.1$ for regression, and from $3.0{-}6.0$ in steps of $0.2$ for classification. The experiments are run for 400 epochs (instead of the 1000 epochs used previously). \autoref{fig:heatmaps} shows the resulting value of $p$ at which the network loss is minimized for each network composition, taking the median across the 10 random seeds.

\autoref{fig:heatmaps} confirms that the optimal regret-to-weight mapping slope $p$ is largely independent of the network composition, further supporting the use of default values of $p=3$ (regression) and $p=5$ (classification), independently of the network composition. The only trend that might exist is one of lower optimal $p$ when $N_{\rm f}>N_{\rm i}$ (towards the bottom-right corner of the heatmap), and higher optimal $p$ when $N_{\rm i}>N_{\rm f}$ (towards the top-left corner of the heatmap). For regression, the statistical significance is questionable, with mean optimal $\overline{p}=2.79\pm0.02$ across all compositions, $\overline{p}=2.81\pm0.03$ when the number of inferers is greatest, and $\overline{p}=2.79\pm0.04$ when the number of forecasters is greatest. For classification, the trend is marginally significant: we find $\overline{p}=4.79\pm0.04$ across all compositions, $\overline{p}=4.85\pm0.05$ when the number of inferers is greatest, and $\overline{p}=4.70\pm0.08$ when the number of forecasters is greatest. This slight trend of increasing optimal $p$ with $N_{\rm i}/N_{\rm f}$ reflects the consistently high performance of forecasters, which means that forecaster-heavy network compositions benefit more from averaging (lower $p$), whereas inferer-heavy network compositions benefit more from inference selection (higher $p$).

\begin{figure}
  \centering
  \includegraphics[width=0.9\linewidth]{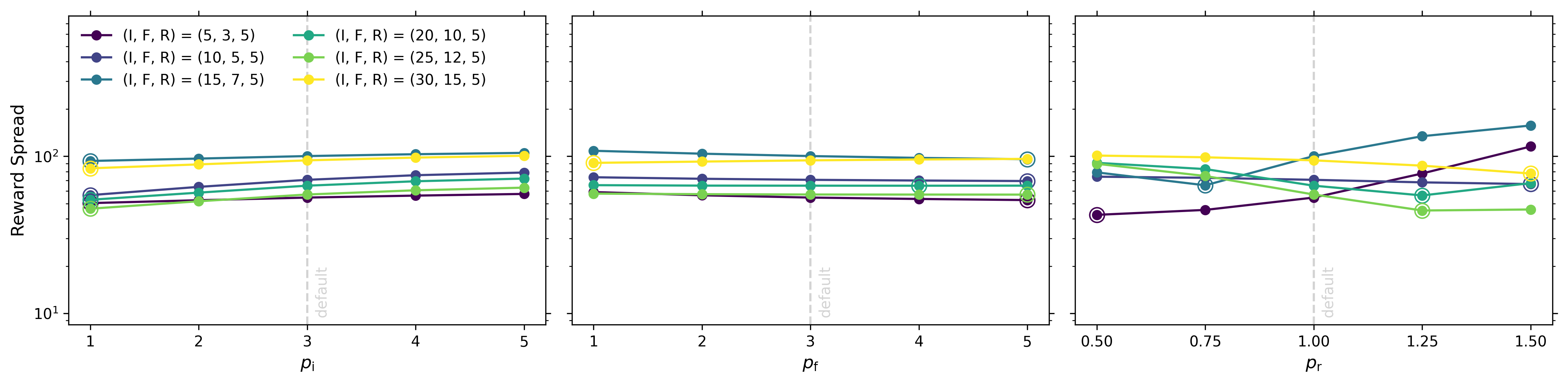}%
  
  \includegraphics[width=0.9\linewidth]{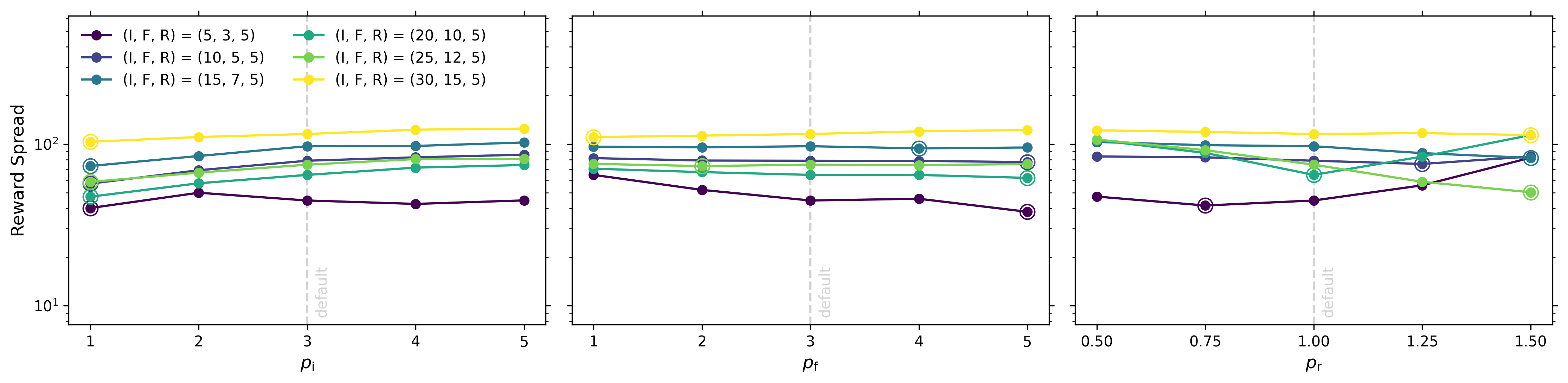}%
  \caption{Variation in median reward spread for different network compositions, as a function of the score-to-reward mapping slope $p_{\rm i}$ (left), $p_{\rm f}$ (middle), and $p_{\rm r}$ (right), for regression (top) and classification (bottom). The median is taken over 1000 epochs, with each line representing a different network composition as indicated by the legend. For each network composition, an open circle marks the $x$-value of the minimum reward spread for that composition. The vertical dashed lines show the default values of $p_{\rm i}=3$, $p_{\rm f}=3$, and $p_{\rm r}=1$.
  }
  \label{fig:reward_spread_composition}
\end{figure}

\autoref{fig:reward_spread_composition} shows the reward spread as a function of the score-to-reward mapping slopes $p_{\rm i}$, $p_{\rm f}$, and $p_{\rm r}$ for the different network compositions. The overall (lack of) trends between the reward spread and the score-to-reward mapping slopes are similar to those shown in \autoref{fig:reward_spread}. As before in the discussion of \autoref{fig:reward_spread}, we see that decreasing the score-to-reward mapping slopes for classes with relatively more numerous participants (characterized by low mean reward per participant) increases their reward entropy, boosts their allocated rewards, and therefore results in a lower reward spread between the classes. Similarly, increasing the score-to-reward mapping slopes for classes with relatively fewer participants (characterized by high mean reward per participant) decreases their reward entropy, reduces their allocated rewards, and therefore also results in a lower reward spread between the classes. This is nicely illustrated by the right-hand panels, where the reward spread increases with $p_{\rm r}$ when reputers are relatively numerous compared to inferers and forecasters (darkest lines), and decreases with $p_{\rm r}$ when reputers are relatively few in number (lighter lines). The same distinction is seen by comparing the general trend between the reward spread and $p_{\rm i}$ (many participants, increasing trend) and $p_{\rm f}$ (few participants, decreasing trend). It depends on the desired incentive structure of a decentralized AI network which parameter setup is preferred, but the trends are generally weak enough to consider this a second-order effect. This conclusion holds in particular, because in practice decentralized intelligence networks will often limit the number of participants in each class of activity (e.g.\ through merit-based sortition schemes, see \citealt{kruijssen24b}), such that the dynamic range of the reward spread can be controlled. In conclusion, the default values of $p_{\rm i}=3$, $p_{\rm f}=3$, and $p_{\rm r}=1$ are able to accommodate a wide range of network compositions.

\section{Discussion and Conclusion} \label{sec:disc}
We present a generalizable framework for optimizing key parameters of a decentralized inference system, with a focus on comparing regression and classification tasks. In order to test this framework, we extend the Inference Synthesis mechanism presented in \citet{kruijssen24} to additionally handle classification tasks. This extension requires minor modifications, but unlocks a wide variety of applications. The ability of Allora's classification solution to handle unbounded label sets provide a flexible basis for further extensions of the protocol to accommodate unsupervised learning tasks (e.g.\ clustering, dimensionality reduction) or even generative tasks (e.g.\ large-language models). Models handling such tasks share some key similarities with the unbounded multi-class classification solution that we have presented here, and we expect future development of the Allora Network to benefit from these parallels.

We describe a comprehensive suite of numerical experiments using synthetic data to systematically assess the impact of the network's key parameters on its performance. Across a broad range of network compositions, we show that there exists an optimal slope for the mapping between performance (here quantified through a regret) and the weight used to express the contribution of an individual inference to the network inference. For the specific use case of the Allora Network, we find optimal slopes of $p=3$ for regression and $p=5$ for classification. The difference arises due to the bounded nature of label probabilities in classification, implying that model selection (i.e.\ the ability to choose the model that predicts the best label, requiring higher $p$) is more important for classification, whereas the unbounded nature of regression favors the use of averaging (lower $p$) to dampen the impact of incorrect inferences.

Likewise, we show that the weight attached to historical performance is optimized when at least 10 inference cycles are included (EMA parameter $\alpha\leq0.1$), and the historical performance is measured within the limits of the full network history ($\alpha>1/N_{\rm epochs}$). Since performance variability is a concern for the accuracy of the network inference, we recommend maximizing network adaptability by setting $\alpha$ to the highest value within these limits, i.e.\ $\alpha=0.1$.

Finally, we optimize the mapping between performance and rewards. For inference-contributing participants (`inferers' and `forecasters' within the context of network design adopted here), performance is quantified through a score that is constructed as a loss difference between the network inference with and without an individual inference. For staking participants that evaluate worker performance (`reputers' within the context of network design adopted here), performance is quantified through a score that is constructed as the product of their stake (monetary commitment) and their proximity to the consensus across the entire reputer pool. To sustain the network's decentralization and the incentivation of the necessary participants, the reward per participant should be similar across the various classes of activity. We achieve this by minimizing the reward spread, i.e.\ the standard deviation of the mean reward received across each class of activity. For both regression and classification, we find that the slopes of the score-to-reward mapping has a minor (but understandable) impact on the reward spread, such that $p_{\rm i}=3$, $p_{\rm f}=3$, and $p_{\rm r}=1$ form a suitable selection of default parameters.

The above results apply nearly independently of network composition. We identify a statistically marginally significant dependence of the optimal regret-to-weight mapping slope $p$ on the network composition (with $p$ increasing with the inferer-to-forecaster ratio), but the dynamic range of the network loss remains sufficiently small to favor the simplicity of a single parameter choice over composition-dependent optimization. The framework presented here thus provides a practical recipe for the calibration of Inference Synthesis and incentive structures in decentralized intelligence networks focused on solving regression and classification tasks. In the future, we expect to extend the framework to the optimization of other decentralized learning and AI tasks.

\begin{sloppypar}
\bibliographystyle{wp}
{\small
\bibliography{ourbib}
}
\end{sloppypar}

\end{document}